\documentclass[runningheads]{llncs}

\usepackage{eccv}

\usepackage{eccvabbrv}
\usepackage{graphicx}
\usepackage{booktabs}
\usepackage{multirow}
\usepackage{amsmath}
\usepackage[accsupp]{axessibility}

\usepackage{hyperref}
\usepackage{orcidlink}

\begin{document}

\title{Emergent Object Binding Has a Finite Spatial Horizon}

\titlerunning{Emergent Object Binding Has a Finite Spatial Horizon}

\author{Mayank Singal}
\authorrunning{M. Singal}
\institute{Independent Researcher}

\maketitle

\begin{abstract}
Pretrained Vision Transformers encode whether two image patches belong to the same object. This \emph{IsSameObject} signal is decodable from frozen patch embeddings at high accuracy, which suggests that object binding emerges from self-supervised pretraining alone. We show that this single accuracy number hides the structure of the signal. Binding is local: the probability that two patches of the same object are decoded as bound falls off monotonically with the distance between them and levels off at a nonzero floor, a falloff well described by an exponential with a finite length scale. This decay holds across object sizes, across three families of probe, on both ADE20K and COCO, and across DINO and CLIP backbones, which indicates that it is a property of the representation rather than of the decoder. Reading binding as local spatial coherence with a finite range accounts for a set of behaviors that the aggregate score leaves unexplained: binding weakens on large objects, separates distinct objects of the same class less reliably than objects of different classes, and groups object parts with their wholes. It is, by contrast, unaffected by occlusion once object size is controlled. We map each behavior with confounds controlled. As a preliminary observation, the horizon and its floor are organized at different depths in DINOv2 and DINOv3, which we report as suggestive given the small number of layers probed and the confound between the two models.
\end{abstract}

\section{Introduction}
\label{sec:intro}

Object binding is the grouping of image features into coherent objects. It is a basic operation of perception, and a recent line of work shows that it emerges in large pretrained Vision Transformers (ViTs) without any object-level supervision~\cite{li2025binding}. The evidence is a probe trained on frozen patch embeddings: given a pair of patches, the probe decodes whether the two patches belong to the same object. On self-supervised models such as DINOv2~\cite{oquab2024dinov2}, this \emph{IsSameObject} probe reaches well above 85\% accuracy, which is read as evidence that the frozen representation already carries object identity.

A single accuracy number tells us that the signal exists. It does not tell us what the signal is. Two patches drawn at random from the same object are usually close together, so a probe can score well by linking nearby patches while never being tested on distant ones. The aggregate number also averages over objects of every size, every class, and every spatial arrangement, which hides any structure that depends on these factors. If the representation held a faithful notion of object identity, the probe would link any two patches of an object equally well, regardless of how far apart they sit in the image, whether a second object of the same class is present, and whether the patches lie on a part or on the whole. Whether the probe does this is the question that the accuracy number cannot answer.

\begin{figure}[t]
  \centering
  \includegraphics[width=\linewidth]{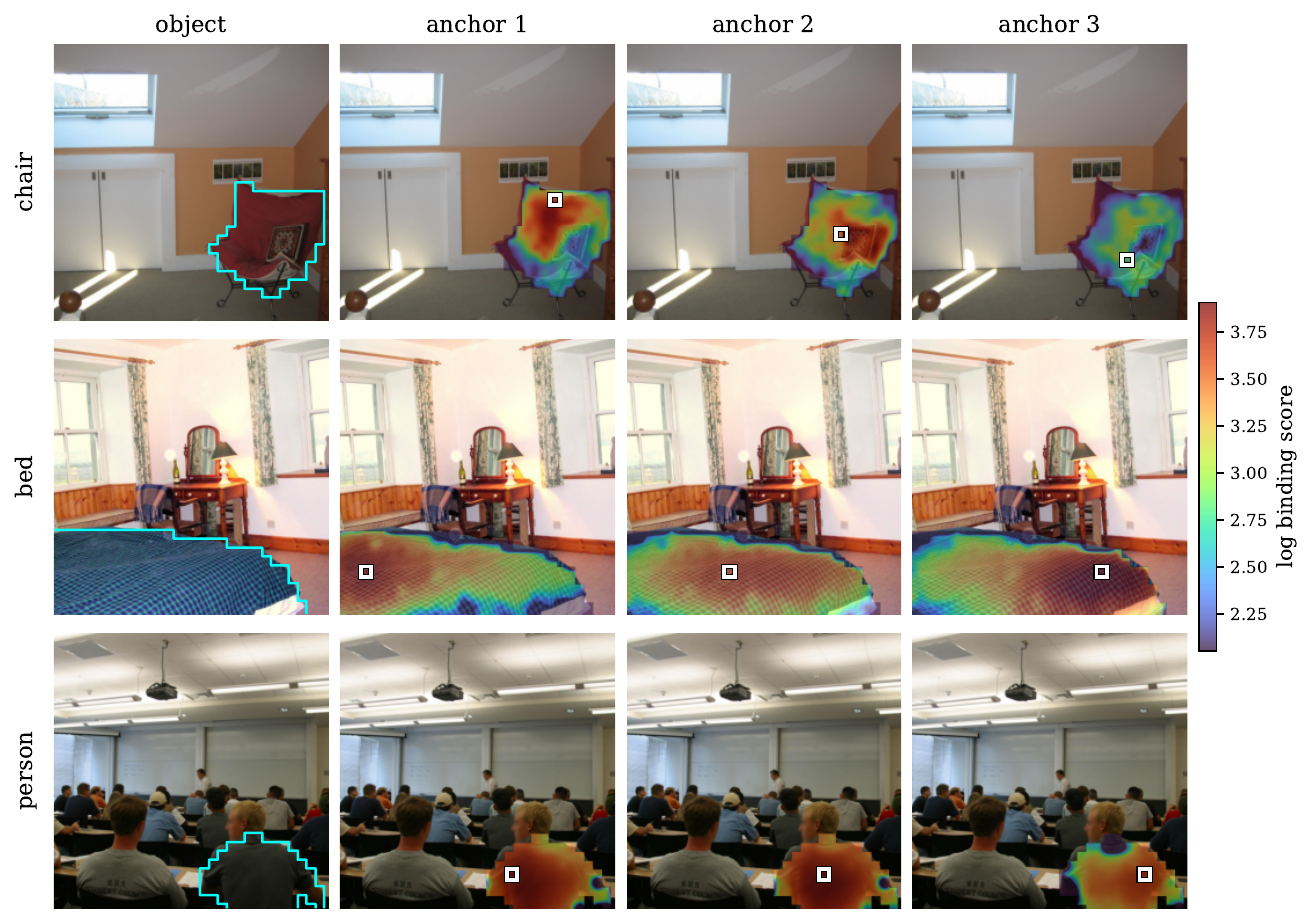}
  \caption{Object binding is spatially local. For each object (left, cyan outline), the panels show the binding score from an anchor patch (white square) to every other patch of the same object. The score peaks at the anchor and falls off with distance, and the bright region follows the anchor as it moves, which illustrates the finite spatial horizon across object types. We show the probe logit, since the decoded probability saturates near one on a clear object at this layer; \cref{fig:horizon} quantifies the decay over the dataset. DINOv2-Large, layer~18.}
  \label{fig:teaser}
\end{figure}

We ask what the IsSameObject signal is, and we find that it is local (\cref{fig:teaser}). We measure how the probability of binding two same-object patches depends on the distance between them, while holding object size fixed. The probability decays with distance, and the decay is well described by an exponential with a finite length scale (\cref{fig:horizon}, \cref{sec:horizon}). The signal links patches that lie within a limited range of each other and weakens steadily beyond it. We call this range the spatial horizon of binding.

The horizon reframes the behavior of the probe. Once binding is read as a local process with a finite range, a set of observations that the aggregate accuracy leaves unexplained follow from a single picture (\cref{sec:failures}). Binding weakens on large objects, because a large object contains patch pairs that exceed the horizon. It separates two distinct objects of the same class less reliably than two objects of different classes, which matches a distance-independent component of the signal that tracks class rather than instance. It groups a part with its whole at the same rate that it groups the whole with itself. It is, in contrast, unaffected by occlusion once object size is controlled.

We establish these results on DINOv2 and DINOv3~\cite{simeoni2025dinov3}, using the dense object, part, and occlusion annotations of ADE20K~\cite{zhou2017ade20k,zhou2019ade20k} and the central decay also on COCO~\cite{lin2014coco}. The spatial horizon is present in both models and is robust to the choice of probe family. We also observe, as a preliminary point, that the two models organize the horizon and its floor at different depths (\cref{sec:depth}); because this rests on few layers and a two-model comparison, we present it as suggestive rather than as a main result.

Our contributions are as follows.
\begin{itemize}
\item We show that the IsSameObject signal in pretrained ViTs decays monotonically with inter-patch distance to a nonzero floor, well described by an exponential with a finite length scale, so binding has a finite spatial horizon. The result holds across object sizes, across linear, bilinear, and MLP probe families, and on both ADE20K and COCO.
\item We show that the horizon accounts for a set of binding behaviors that the aggregate accuracy hides, and we map each on ADE20K with the relevant confounds controlled by matching on distance and on object size and by comparison against a feature-similarity baseline.
\end{itemize}
As a preliminary observation, we also find that the horizon and its semantic floor are organized at different depths in DINOv2 and DINOv3 (\cref{sec:depth}); we report this as suggestive rather than as a primary result, for the reasons given in that section.

The work reuses the probing framework of Li \etal~\cite{li2025binding} and contributes an analysis of what the probed signal is. The reading it supports is that emergent binding behaves as local spatial coherence with a finite range, together with a weaker distance-independent component that tracks semantic class.

\section{Related Work}
\label{sec:related}

\paragraph{Emergent structure in self-supervised features.}
Self-supervised ViTs produce dense features that support segmentation, correspondence, and retrieval without task-specific training~\cite{caron2021dino,oquab2024dinov2,simeoni2025dinov3}. DINO and its successors are known to expose object boundaries in attention maps and to give patch features that cluster by object. DINOv3 extends this line with rotary position embeddings, register tokens, and gated feed-forward blocks, and reports strong dense-prediction performance across benchmarks~\cite{simeoni2025dinov3}. These works study what the features are useful for. We study one specific relational property of the features, namely whether two patches are encoded as belonging to the same object, and how that property is structured.

\paragraph{Object binding.}
Binding refers to the grouping of features into objects, a question with a long history in perception and in slot-based and object-centric models that learn explicit object representations~\cite{locatello2020slot,greff2020binding}. Li \etal~\cite{li2025binding} show that binding need not be built in: an IsSameObject relation is linearly decodable from frozen ViT patch embeddings, most strongly in DINO-family models and weakly in masked-autoencoder models, which suggests that binding is acquired under particular pretraining objectives. They establish that the signal exists and compare it across pretraining methods. We take their probe as a measurement instrument and ask what the decoded signal is: how it depends on spatial distance and where it succeeds and fails.

\paragraph{Probing and its limits.}
Linear and nonlinear probes are a standard tool for testing what a representation encodes~\cite{alain2017probing,belinkov2022probing}. A known difficulty is that a sufficiently expressive probe can fit structure that the representation does not make readily available, so a probed effect may reflect the decoder rather than the features. We address this directly by repeating each central measurement across three probe families of differing capacity and by comparing against a feature-similarity baseline, and we report which effects are stable across probes and which are not.

\section{Method}
\label{sec:method}

\paragraph{The IsSameObject probe.}
We follow the framework of Li \etal~\cite{li2025binding}. Given a frozen ViT, we extract patch embeddings from a chosen layer. For a pair of patch embeddings $\mathbf{x},\mathbf{y}\in\mathbb{R}^{D}$, a probe produces a score $s(\mathbf{x},\mathbf{y})$, trained with a binary cross-entropy loss to predict whether the two patches belong to the same object instance. Our default probe is the bilinear form
\begin{equation}
s(\mathbf{x},\mathbf{y}) = \mathbf{x}^{\top}\mathbf{W}\mathbf{y} + b ,
\label{eq:bilinear}
\end{equation}
with $\mathbf{W}$ symmetrized and scaled by $1/\sqrt{D}$. To test whether our findings depend on this choice, we also use a linear probe on the elementwise product, $s(\mathbf{x},\mathbf{y})=\mathbf{w}^{\top}(\mathbf{x}\odot\mathbf{y})+b$, and a two-layer network on the concatenation of the product and absolute difference, $s(\mathbf{x},\mathbf{y})=g([\mathbf{x}\odot\mathbf{y};\,|\mathbf{x}-\mathbf{y}|])$.

\paragraph{Models and data.}
We study DINOv2-Large and DINOv3-Large, both ViTs with 24 transformer blocks. DINOv3 prepends a class token and four register tokens, which we discard, keeping the patch tokens. We use ADE20K~\cite{zhou2017ade20k,zhou2019ade20k}, which provides per-object instance masks, object classes, a part hierarchy that links parts to their parent objects, and occlusion flags. We map each annotation to the patch grid by majority vote within each patch, and we use the 2000-image validation split. We train probes on patches sampled from a held-out set of images and evaluate on a disjoint set. To test dataset generality we repeat the central distance measurement on COCO instance masks~\cite{lin2014coco}, mapped to the patch grid in the same way.

\paragraph{Pair sampling and metric.}
For training, we sample balanced pairs per image: positive pairs are two patches of the same object instance, and negative pairs are two patches of different instances. Random pairs are dominated by the negative class, so balancing is necessary to avoid a trivial majority-class solution. We train for 35 epochs with Adam and report results over three seeds. Because a fixed decision threshold can favor one class, we report area under the ROC curve where a single summary is needed, and we report class-resolved accuracies elsewhere.

\paragraph{Confound control.}
Object size, inter-patch distance, and object class are correlated in natural images, so a raw effect may reflect a confounder rather than the factor of interest. We control for this by stratified matching: to test factor $A$ while removing factor $B$, we bin pairs by $B$, measure the effect of $A$ within each bin, and report the weighted average across bins. We apply this to remove object size when measuring distance, to remove distance and size when measuring the same-class effect, and to remove size when measuring occlusion.

\section{Binding Has a Finite Spatial Horizon}
\label{sec:horizon}

\paragraph{Setup.}
We restrict attention to positive pairs, two patches of the same object, and ask how the probe's binding probability depends on the Euclidean distance $d$ between the two patches on the patch grid. To remove object size as a factor, we measure within a fixed object-size band, so that only the separation between the two patches varies.

\paragraph{Binding decays with distance.}
The binding probability falls off with distance. \Cref{tab:horizon} reports the binned probability for DINOv3 within a fixed size band. A short separation gives a high binding probability; the probability declines steadily as separation grows and levels off at a distance-independent floor. The decay is well described by
\begin{equation}
P(\text{same}\mid d) = a\,e^{-d/\lambda} + c ,
\label{eq:decay}
\end{equation}
with a coefficient of determination above $0.9$. We summarize the range with the length scale $\lambda$ and report it as a range rather than a single value, because the fitted $\lambda$ is sensitive to the binning and to the probe; the decay itself is stable.

\begin{figure}[t]
  \centering
  \includegraphics[width=0.74\linewidth]{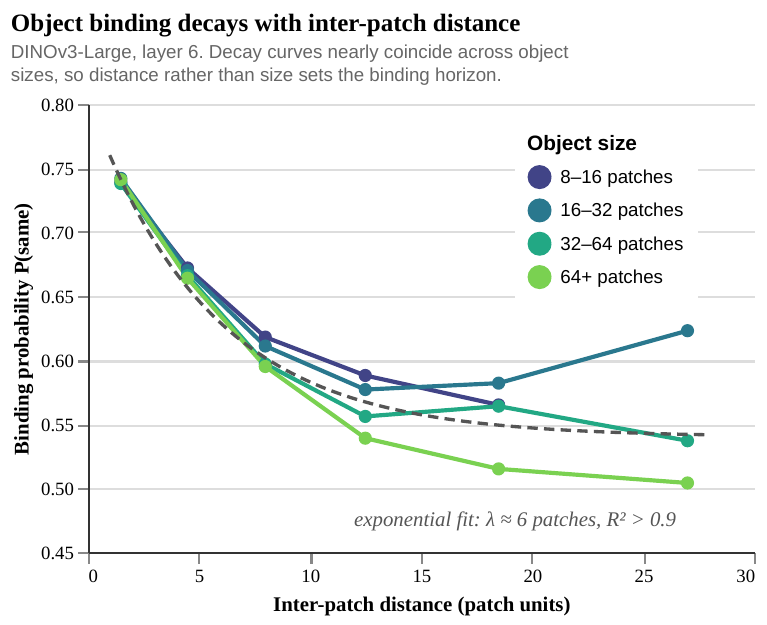}
  \caption{Binding probability against inter-patch distance for same-object pairs, with object size held within a band. Each curve fixes a size band and varies only the separation. The curves nearly coincide, so distance rather than object size sets the range over which binding holds. The dashed line is an exponential fit pooled across bands. Bins beyond about 17 patches are sparse and are omitted. DINOv3-Large, layer~6.}
  \label{fig:horizon}
\end{figure}

\paragraph{The decay is governed by distance, not size.}
Repeating the measurement across size bands gives curves that lie close to one another (\cref{tab:horizon}, \cref{fig:horizon}). The dependence of binding on object size, which a measurement that ignores distance recovers, is to first order a consequence of larger objects admitting more distant patch pairs. Distance is the variable that organizes the signal.

\begin{figure}[t]
  \centering
  \includegraphics[width=0.78\linewidth]{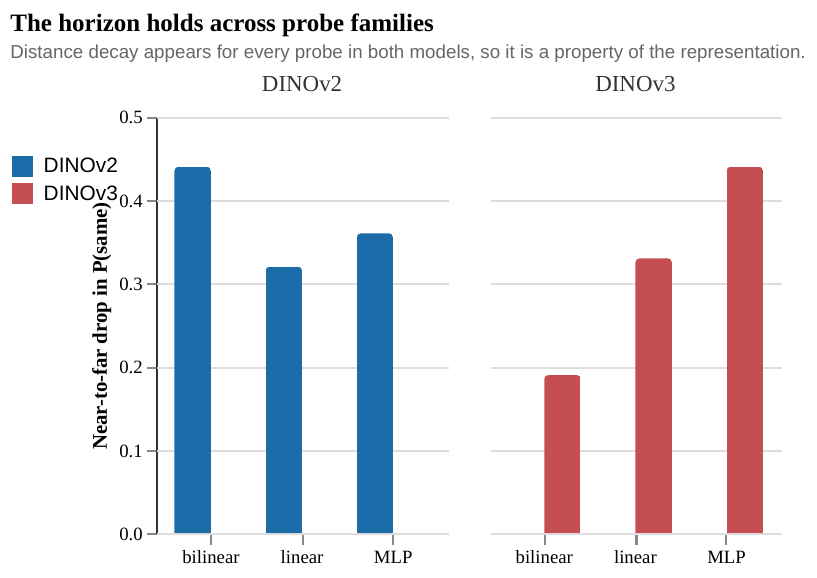}
  \caption{The distance decay appears under every probe family in both models. ``Near-to-far drop'' is the fall in binding probability from near pairs ($d<4$) to far pairs ($d>15$) within a fixed object-size band. Because the decay does not depend on the choice of decoder, it reflects the representation rather than the probe.}
  \label{fig:probes}
\end{figure}

\paragraph{The decay is a property of the representation.}
We repeat the measurement with all three probe families. \Cref{tab:probes} and \cref{fig:probes} report the near-to-far drop in binding probability and the fit quality for each. The exponential decay appears for the bilinear, linear, and MLP probes alike, with high fit quality, in both DINOv2 and DINOv3. The magnitude of the drop does depend on the decoder: in DINOv3 the near-to-far drop ranges from $0.19$ to $0.44$ across the three probes. What is invariant is the presence and the sign of the decay, not its size. We therefore read the existence of a finite horizon, rather than any particular length scale, as the property of the representation, which is consistent with reporting $\lambda$ as a range.

\paragraph{The decay replicates on a second dataset.}
To test whether the horizon is specific to ADE20K, we repeat the core measurement on COCO. We map COCO instance masks to the patch grid by the same majority vote, train the probe on a disjoint set of COCO images, and fit \cref{eq:decay} on held-out COCO images, using DINOv3 at the same early layer. The decay reproduces: binding probability falls monotonically with distance from $0.69$ at the nearest bin to $0.44$ beyond fifteen patches, with a finite length scale of $\lambda\!\approx\!5.5$, a nonzero floor of $c\!\approx\!0.43$, and a fit quality of $R^2=0.97$. Both terms of \cref{eq:decay} are present, with a length scale and a floor of the same order as on ADE20K. The finite horizon and the semantic floor are therefore not artifacts of one dataset.

\paragraph{The horizon is not specific to DINO.}
To test whether the horizon is a property of the DINO lineage or of pretrained ViTs more broadly, we repeat the central measurement on two non-DINO ViT-L models on COCO: CLIP~\cite{radford2021clip}, trained with a contrastive image-text objective, and MAE~\cite{he2022mae}, trained by masked reconstruction. The result depends on whether the binding signal is present at all. In CLIP the IsSameObject signal is decodable at strength comparable to DINO (probe AUC $0.85$), and it shows the same structure: a monotonic decay to a nonzero floor, well fit by \cref{eq:decay} ($\lambda\!\approx\!7.6$, $c\!\approx\!0.42$, $R^2=0.95$). In MAE the binding signal is near chance (AUC $0.54$), so there is no binding structure to decay; this reproduces the weak binding that Li \etal~\cite{li2025binding} report for masked-autoencoder models. The horizon and its floor therefore appear wherever the IsSameObject signal is present, across both self-distillation and contrastive pretraining, and are absent where the signal itself is weak (\cref{tab:models}). This locates the phenomenon in the binding signal rather than in any single architecture or pretraining method. These measurements use one layer and one probe per model and are a targeted check rather than a full cross-architecture study.

\begin{table}[tb]
  \caption{The horizon across pretraining objectives, measured on COCO. AUC is the strength of the IsSameObject signal; the horizon is read from \cref{eq:decay}. The decay appears wherever the signal is present and is absent where it is near chance. ViT-L backbones; one layer and one probe per model.}
  \label{tab:models}
  \centering
  \begin{tabular}{@{}llcc@{}}
    \toprule
    Model & Pretraining & Probe AUC & Horizon\\
    \midrule
    DINOv3-L & self-distillation        & ${\sim}0.88$ & present ($\lambda\!\approx\!5.5$, $c\!\approx\!0.43$)\\
    CLIP-L   & contrastive image-text   & $0.85$       & present ($\lambda\!\approx\!7.6$, $c\!\approx\!0.42$)\\
    MAE-L    & masked reconstruction    & $0.54$       & absent (signal near chance)\\
  \bottomrule
  \end{tabular}
\end{table}

\begin{table}[tb]
  \caption{Binding probability $P(\text{same}\mid d)$ for same-object patch pairs in DINOv3, by inter-patch distance and object-size band. The decay with distance is present at every size, and the curves are close across sizes, which shows that distance rather than size organizes the signal. Distances are in patch units. A dash marks a cell where the object-size band cannot contain pairs at that separation. The farthest bins are sparsely populated, so small non-monotonic fluctuations there (for example the final cell of the 16--32 row) reflect bin noise rather than a reversal of the decay.}
  \label{tab:horizon}
  \centering
  \begin{tabular}{@{}lcccccc@{}}
    \toprule
    Object size & $0\text{--}3$ & $3\text{--}6$ & $6\text{--}10$ & $10\text{--}15$ & $15\text{--}22$ & $22\text{--}32$\\
    \midrule
    8--16   & 0.74 & 0.67 & 0.62 & 0.59 & 0.57 & --   \\
    16--32  & 0.74 & 0.67 & 0.61 & 0.58 & 0.58 & 0.62 \\
    32--64  & 0.74 & 0.67 & 0.60 & 0.56 & 0.56 & 0.54 \\
    64+     & 0.74 & 0.66 & 0.60 & 0.54 & 0.52 & 0.50 \\
  \bottomrule
  \end{tabular}
\end{table}

\begin{table}[tb]
  \caption{The spatial decay across probe families, measured within a fixed object-size band. ``Drop'' is the fall in binding probability from near pairs ($d<4$) to far pairs ($d>15$); $R^2$ is the fit quality of \cref{eq:decay}. The decay is present for every probe family in both models, which indicates a property of the representation rather than of the decoder. DINOv3 is measured at its early binding layer and DINOv2 at its late binding layer (\cref{sec:depth}).}
  \label{tab:probes}
  \centering
  \begin{tabular}{@{}llcc@{}}
    \toprule
    Model & Probe & Near-to-far drop & $R^2$\\
    \midrule
    DINOv3 & bilinear   & 0.19 & 0.92\\
    DINOv3 & linear      & 0.33 & 0.97\\
    DINOv3 & MLP         & 0.44 & 0.98\\
    \midrule
    DINOv2 & bilinear   & 0.44 & 0.97\\
    DINOv2 & linear      & 0.32 & 0.96\\
    DINOv2 & MLP         & 0.36 & 0.97\\
  \bottomrule
  \end{tabular}
\end{table}

\paragraph{Interpretation.}
The representation links patches that lie within a limited distance of each other and links them weakly beyond it. \Cref{eq:decay} separates the signal into two parts: a distance-dependent term $a\,e^{-d/\lambda}$ with a finite range, and a distance-independent floor $c$. The local term is the spatial horizon. The floor is a residual binding signal that does not decay with distance, and \cref{sec:failures} shows that it tracks object class. Binding therefore behaves as local spatial coherence with a finite range, together with a weaker semantic component.

\paragraph{What the result is and is not.}
A decay of binding with distance is, on its own, close to expected. Patch features vary smoothly over an image, nearby patches have correlated features, and any decoder that reads binding from feature overlap will report higher binding for nearby patches. We do not claim the local term as a novel mechanism, and we do not read the exponential form as evidence of anything beyond a smooth falloff. The contribution is the structure that feature proximity alone does not predict. First, the signal does not decay to chance: it levels off at a floor $c$ that, at some layers, is far above chance, so distant same-object patches remain bound at a rate that proximity cannot explain. Second, this floor tracks object class rather than distance, and the resulting same-class effect exceeds a feature-similarity baseline by about a factor of two (\cref{sec:failures}); a thresholded cosine-similarity classifier, which is exactly the ``nearby features are similar'' account made explicit, recovers less than half of the effect that the trained probe shows. The probe therefore reads out a binding relation that raw feature overlap does not. Third, both the floor and its same-class correlate reproduce on COCO (\cref{sec:horizon,sec:failures}), so they are not specific to one dataset. We treat the local term as the expected part and rest the analysis on the floor, the class dependence above the similarity baseline, and the replication across datasets.

\section{The Horizon Accounts for Where Binding Breaks}
\label{sec:failures}

Reading binding as a local process with a finite range and a semantic floor predicts where it should succeed and where it should not. We test each prediction on ADE20K, controlling the confounds that \cref{sec:horizon} exposes.

\paragraph{Large objects.}
Binding weakens as objects grow. Measured on same-object pairs, the binding accuracy falls from $97.5\%$ on small objects (2--4 patches) to $61.6\%$ on the largest (64+ patches), monotonically across size. This follows from the horizon: a large object contains patch pairs whose separation exceeds the binding range, so the representation does not hold the whole object together. This is the size dependence of \cref{sec:horizon} viewed from the other side, and \cref{sec:horizon} shows that the underlying variable is distance.

\paragraph{Same-class instances.}
When two patches belong to two different objects of the same class, the probe rejects them as different less often than it does for two objects of different classes. At each model's peak layer the gap is $12$ to $13$ points ($13.5$ in DINOv2, $12.5$ in DINOv3). It is stable across three seeds. It survives matching on inter-patch distance, where it remains $12.6$ points, and on object size, where it remains $13.1$ points, so it is not an artifact of either confounder; these matched figures are pooled over the peak-layer measurement and are not tied to a single depth in \cref{tab:floor}. It also exceeds a feature-similarity baseline: a cosine-similarity classifier with a threshold tuned for balanced accuracy shows a same-class gap of $4.8$ points in DINOv2 and $6.0$ points in DINOv3, while the trained probe shows $12$ to $13$ points, roughly twice the similarity baseline. The effect is therefore a property of the binding representation beyond raw feature overlap. The same deficit appears on COCO, where the probe reaches a lower binding accuracy on the more cluttered scenes: the size-matched same-class gap is $10.5$ points, close to the ADE20K value, and within objects small enough to sit inside the horizon (under $24$ patches) it rises to $27.8$ points. The deficit is thus concentrated where binding is strong and fades on large objects, which is what the horizon predicts: beyond the spatial range there is little binding of any kind to bias toward class. It is the correlate of the distance-independent floor in \cref{eq:decay}: within the spatial horizon, the residual binding signal groups patches by class rather than by instance.

\paragraph{Parts and wholes.}
We use the ADE20K part hierarchy to form pairs in which one patch lies on a part, such as a handle, and the other on its parent whole, such as a mug. The probe groups the part with its whole at the same rate that it groups two patches within the whole, and far above the rate for two unrelated objects. The representation groups a part with its whole rather than treating the part as a separate object.

\paragraph{Occlusion.}
Occluded objects appear to bind better than unoccluded ones, by about $8$--$10$ points. This effect is a size confound: occluded objects in ADE20K tend to be larger and more central. After matching on object size, the effect falls to about a point of zero or less in both models ($0.3\pm0.7$ in DINOv3, $-1.3\pm0.5$ in DINOv2). Binding is unaffected by occlusion once size is controlled. This null result also serves as a check on the method: the same matching procedure that removes the occlusion effect leaves the same-class effect intact, which argues that the same-class effect is not a generic artifact of the pipeline.

\section{A Preliminary Note on Depth}
\label{sec:depth}

This section is exploratory and is not one of our main claims. It probes only four layers per model, the exponential fit is unreliable at the intermediate layers, and any DINOv2-versus-DINOv3 difference confounds the depth of binding with every other difference between the two models. We report the pattern as suggestive and draw no causal conclusion from it.

\begin{figure}[t]
  \centering
  \includegraphics[width=0.72\linewidth]{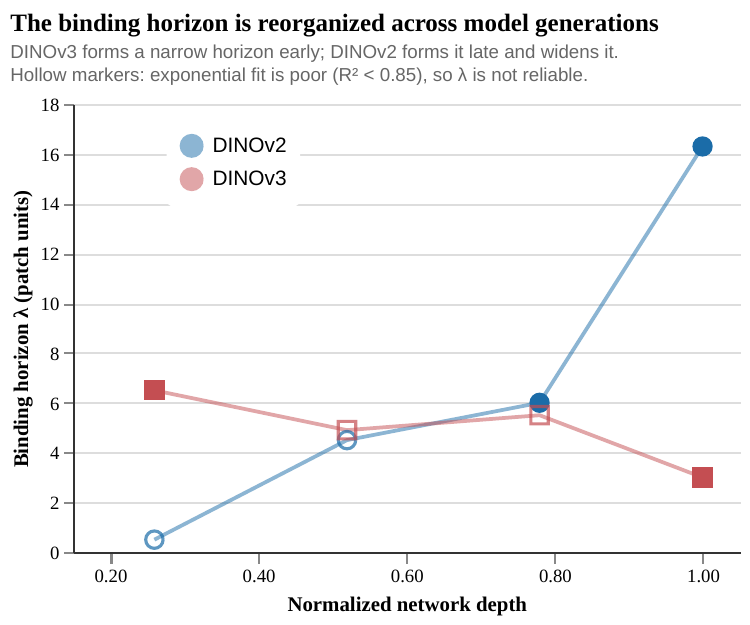}
  \caption{The fitted binding horizon $\lambda$ across network depth. DINOv3 forms a clean narrow horizon early, while DINOv2 forms its horizon late and widens it toward the output. Hollow markers indicate layers where the exponential fit is poor ($R^2<0.85$), so the corresponding $\lambda$ is not reliable; the early-versus-late contrast and the fit quality, rather than any single value, are the robust observation.}
  \label{fig:depth}
\end{figure}

The spatial horizon is present in both DINOv2 and DINOv3, and the two models appear to build it at different depths. We fit \cref{eq:decay} at four layers spanning each network and read the fit quality and the length scale (\cref{fig:depth}). In DINOv3 the horizon is well formed early, within the first quarter of the network, with a clean exponential fit at a short length scale. In DINOv2 the horizon is not well described by an exponential at mid-depth and becomes clean only in the final third of the network. The fitted length scale at the intermediate layers is unreliable in both models, and the trajectory is noisy, so we read the pattern from the endpoints and from the fit quality rather than from the fitted values. The same-class effect shows a similar split: it is largest early in DINOv3 and at mid-depth in DINOv2. \Cref{tab:depth} reports the same-class gap across depth for both models.

\begin{table}[tb]
  \caption{Same-class gap (in points, size-matched) across network depth, for DINOv2 and DINOv3, over three seeds. The effect peaks at mid-depth in DINOv2 and early in DINOv3, which shows that the behavior is shared between the generations while its location in the network differs. Depth is the layer index normalized to $[0,1]$.}
  \label{tab:depth}
  \centering
  \begin{tabular}{@{}lcccc@{}}
    \toprule
    Depth & $0.26$ & $0.52$ & $0.78$ & $1.00$\\
    \midrule
    DINOv2 & $6.7\pm2.5$ & $13.5\pm0.5$ & $6.8\pm0.2$ & $4.0\pm2.1$\\
    DINOv3 & $12.5\pm2.1$ & $6.4\pm0.9$ & $5.0\pm1.1$ & $5.6\pm2.1$\\
  \bottomrule
  \end{tabular}
\end{table}

The two models differ in architecture: DINOv3 adds rotary position embeddings, register tokens, and gated feed-forward blocks. Any of these, alone or in combination, could account for the depth difference, and our two-model comparison cannot separate them. We note the association and leave its cause open.

\paragraph{The floor and the semantic effect co-vary across depth.}
The floor $c$ in \cref{eq:decay} is the distance-independent part of the signal, and \cref{sec:failures} shows that the same-class effect is its behavioral correlate. If the two reflect a common semantic component, they should rise and fall together across depth. \Cref{tab:floor} reports $c$, the length scale $\lambda$, and the size-matched same-class gap at four layers for both models. The two move together across depth. In DINOv3 the peaks coincide: the floor and the same-class gap are both largest at the earliest layer we probe. In DINOv2 the floor is largest early (depth $0.26$) while the same-class gap peaks one layer later (depth $0.52$), so the peaks are adjacent rather than identical, but the two quantities still rise and fall together over the four layers. The floor and the gap are positively correlated across the four layers in each model, but with only four points per model we treat this as illustrative and not as a statistical result. In DINOv2 the floor and the length scale move in opposite directions: the floor falls monotonically from $0.53$ to near zero at the output while the length scale grows, so binding shifts from a semantic regime to a geometric one through the network. DINOv3 reaches a low floor by mid-network, which places its semantic component early. These observations are consistent with the floor being the semantic component of binding rather than a fit nuisance, and with its depth profile tracking where same-class instances are over-merged.

\begin{table}[tb]
  \caption{The floor $c$, length scale $\lambda$, and size-matched same-class gap (points) across depth, over three seeds. The floor and the gap track each other across depth: their peaks coincide in DINOv3 (both at depth $0.26$) and are adjacent in DINOv2 (floor at $0.26$, gap at $0.52$). With four layers per model this co-variation is illustrative rather than statistical. Depth is the layer index normalized to $[0,1]$. The fitted $\lambda$ is sensitive to binning and to the run and is included to show the trade-off with the floor, not as a precise value; \cref{fig:depth} gives the trajectory. Values of $\lambda$ here are separate fits from those in \cref{fig:depth} and may differ at intermediate layers.}
  \label{tab:floor}
  \centering
  \begin{tabular}{@{}llcccc@{}}
    \toprule
    Model & Quantity & $0.26$ & $0.52$ & $0.78$ & $1.00$\\
    \midrule
    \multirow{3}{*}{DINOv2}
      & floor $c$            & $0.53$ & $0.50$ & $0.30$ & $0.01$\\
      & $\lambda$            & $7.1$  & $9.6$  & $11.1$ & $15.7$\\
      & same-class gap       & $6.7$  & $13.5$ & $6.8$  & $4.0$\\
    \midrule
    \multirow{3}{*}{DINOv3}
      & floor $c$            & $0.48$ & $0.07$ & $0.12$ & $0.16$\\
      & $\lambda$            & $5.8$  & $15.0$ & $14.1$ & $6.5$\\
      & same-class gap       & $12.5$ & $6.4$  & $5.0$  & $5.6$\\
  \bottomrule
  \end{tabular}
\end{table}

\section{Limitations}
\label{sec:limitations}

We measure distance on the patch grid with Euclidean separation, which approximates but does not equal geodesic distance along an object. For irregular or elongated shapes the two differ, so the reported length scale is best read as an upper bound on the within-object range, and the exact form of the decay for such shapes is left open. The fitted length scale is sensitive to the binning and to the probe family, so we report it as a range; the presence of the decay and its independence from size are the robust claims. The depth analysis uses four layers per model and so locates the horizon coarsely. On scope, the central distance decay reproduces on COCO as well as ADE20K and across DINO, CLIP, and MAE backbones (\cref{sec:horizon}), and the same-class deficit reproduces on COCO (\cref{sec:failures}), so the horizon and its semantic correlate are not artifacts of one dataset or of the DINO lineage. The cross-model check uses one layer and one probe per model and is targeted rather than exhaustive; a full sweep over layers, probes, and more architectures (including SigLIP and supervised ViTs) would sharpen the picture. The part-whole and occlusion analyses, which require part hierarchies and occlusion flags that COCO does not provide, and the depth study, are carried out on ADE20K and the DINO models alone. Finally, we establish the spatial law empirically and do not isolate its cause. Receptive-field geometry, attention locality, positional encoding, and the statistics of object masks in the training data are candidate explanations that the present evidence does not separate, and we do not claim a mechanism. A targeted intervention, such as ablating or perturbing positional information, is the natural next step toward a causal account.

\section{Conclusion}
\label{sec:conclusion}

The IsSameObject signal in pretrained Vision Transformers, which a single accuracy number presents as near-complete, is spatially local. The probability that two patches of the same object are bound decays monotonically with their separation to a nonzero floor, well described by an exponential with a finite range, independent of object size, across three probe families and on both ADE20K and COCO. A weaker component of the signal does not decay with distance and tracks object class. This account explains a set of behaviors that the aggregate score hides: binding weakens on large objects, groups distinct same-class objects more readily than different-class objects, and groups parts with their wholes, while remaining robust to occlusion. Read this way, emergent binding is local spatial coherence with a finite range together with a semantic component, and a single accuracy figure does not capture it. The structure is not specific to the DINO objective: it appears in CLIP with comparable strength and is absent in MAE, where the binding signal is itself near chance, so it tracks the binding signal rather than any one architecture or pretraining method. The clearest remaining direction is a causal account, through an intervention on positional information, and a fuller sweep across layers, probes, and architectures to map where the signal does and does not emerge.

\section*{Use of AI Tools}
The authors used AI-based assistants during this work for manuscript drafting and editing, for writing and debugging analysis code, and for assistance with the experimental analysis. All research claims, experimental design decisions, and final interpretations are the authors' own.


\bibliographystyle{splncs04}
\bibliography{main}
\end{document}